\documentclass[journal]{IEEEtran}

\usepackage{xcolor,soul,framed} %,caption
\usepackage[nolist]{acronym}
\usepackage{float}
\usepackage{subfig}
\usepackage{algorithm}
\usepackage{algpseudocode}
\usepackage{bbm}
\usepackage{comment}
\makeatletter
\newcommand\fs@norules{\def\@fs@cfont{\bfseries}\let\@fs@capt\floatc@ruled
  \def\@fs@pre{}%
  \def\@fs@post{}%
  \def\@fs@mid{\kern3pt}%
  \let\@fs@iftopcapt\iftrue}
\makeatother
\colorlet{shadecolor}{yellow}
\usepackage[pdftex]{graphicx}
\graphicspath{{../pdf/}{../jpeg/}}
\DeclareGraphicsExtensions{.pdf,.jpeg,.png}

\usepackage[cmex10]{amsmath}
\usepackage{array}
\usepackage{mdwmath}
\usepackage{mdwtab}
\usepackage{eqparbox}
\usepackage{url}
\usepackage[colorlinks=true,urlcolor=blue]{hyperref} 
\usepackage{amsfonts}
\usepackage{cleveref}
\usepackage{subfig}
\usepackage[scr=rsfs]{mathalpha}
\usepackage{makecell}
\usepackage{float} 
\usepackage[utf8]{inputenc}
\usepackage{booktabs} % To thicken table lines
   
\usepackage{mathtools}
\usepackage{array}
\newcolumntype{P}[1]{>{\centering\arraybackslash}p{#1}}
\newcolumntype{M}[1]{>{\centering\arraybackslash}m{#1}}

\newcommand{\bX}{\textbf{X}}

\crefalias{subequation}{equation} 
\crefalias{eqnarray}{equation} 
\crefformat{pluraleq}{(#2#1#3)}

\usepackage{bm}

\newcommand{\rev}{\textcolor{black}}

\DeclareUnicodeCharacter{2212}{-}
\begin{document}
\bstctlcite{IEEEexample:BSTcontrol}
    %\title{f wildfire burned areas in a  sequences of burned area for wildfire events: applied to efficient surrogate modelling for fire spread prediction}
    \title{Deep learning surrogate models of JULES-INFERNO for wildfire prediction on a global scale}
\author{Sibo Cheng, Hector Chassagnon,  Matthew Kasoar, Yike Guo \textit{IEEE Fellow}, and Rossella Arcucci
\thanks{corresponding: sibo.cheng@enpc.fr}    

\thanks{Sibo Cheng is with CEREA, \'{E}cole des Ponts and EDF R\&D, \^Ile-de-France, France. }
\thanks{Rossella Arcucci and Hector Chassagnon are with Department of Earth Science and Engineering, Imperial College London, SW7 2AZ, UK. }
\thanks{ Matthew Kasoar is with Department of physics, Imperial College London, SW7 2AZ, UK. }
\thanks{Yike Guo is with Department of Computer Science and Engineering, The Hong Kong University of Science and Technology , 999077, Hongkong, China. }

}

% The paper headers
\markboth{submitted to IEEE Transactions on Emerging Topics in Computational Intelligence,
2023}{Cheng {\textit{et al.}}: Spectral Cross-Domain Neural Network with Soft-adaptive Threshold Spectral
Enhancement}

% ====================================================================
\maketitle

% === ABSTRACT ====================================================================
% =================================================================================
\begin{abstract}
Global wildfire models play a crucial role in anticipating and responding to changing wildfire regimes. JULES-INFERNO is a global vegetation and fire model simulating wildfire emissions and area burnt on a global scale. However, because of the high data dimensionality and system complexity, JULES-INFERNO's computational costs make it challenging to apply to fire risk forecasting with unseen initial conditions. Typically, running JULES-INFERNO for 30 years of prediction will take several hours on High Performance Computing (HPC) clusters. To tackle this bottleneck, two data-driven models are built in this work based on Deep Learning techniques to surrogate the JULES-INFERNO model and speed up global wildfire forecasting.
More precisely, these machine learning models take global temperature, vegetation density, soil moisture and previous forecasts as inputs to predict the subsequent global area burnt on an iterative basis.
 Average Error per Pixel (AEP) and Structural Similarity Index Measure (SSIM) are used as metrics to evaluate the performance of the proposed surrogate models. A fine tuning strategy is also proposed in this work to improve the algorithm performance for unseen scenarios.
Numerical results show a strong performance of the proposed models, in terms of both computational efficiency (less than 20 seconds for 30 years of prediction on a laptop CPU) and prediction accuracy (with AEP under 0.3\% and SSIM over 98\% compared to the outputs of JULES-INFERNO). The codes that were used for building and testing the surrogate models using Python language (3.7) are available at 
\href{https://github.com/DL-WG/DL_Surrogate_JULES_INFERNO}{github}.
\end{abstract}

%\section*{Highlights}
%\begin{itemize}
%    \item Two deep learning-based surrogate models of JULES-INFERNO are created for long-term global wildfire prediction.
%    \item A fine tuning strategy is proposed for predicting unseen scenarios with considerably different initial conditions.  
%    \item We investigated the impact that including different predictor variables has on our surrogate model's skill through extensive numerical experiments.
%    \item The online computational cost can be substantially reduced by the surrogate models (several seconds on a laptop CPU) compared to the JULES-INFERNO system (5 hours on a 32 threads HPC).
%\end{itemize}

%\section*{Software and data availability}
%The codes that were used for building and testing the surrogate models using Python language (3.7) are available at 
%\url{https://github.com/DL-WG/DL\_Surrogate\_JULES\_INFERNO}. Authors' experimental environment is as follows
%\begin{itemize}
%    \item OS: Google colab platform
%    \item GPU: NVIDIA Tesla P100
%    \item RAM: 16GB
%\end{itemize}

\section{Introduction} \label{INTRO}

\indent Long-term prediction of wildfire at a global scale has been a long-standing challenge. Shorter intense wet seasons and longer hot seasons increased wildfire intensity and frequency, costing billions to governments~\cite{xu_2020_climate, wang_2020_economic}. According to~\cite{doerr_2016_global}, Canada and European countries\footnote{Greece, France, Italy, Portugal and Spain} spent respectively US\$531 million and €2.5 billion annually in wildfire prevention, detection or suppression.

Thus, advanced systems like wildfire models, capable of giving robust and accurate predictions of wildfires activities, have revealed themselves as keys to preventing, detecting or managing changing fire risk. Wildfires models that can forecast fire propagation~\cite{finney_1998_farsite}, contribute to alleviating damages, managing firefighting resources or identifying at-risk areas to defend or evacuate. \rev{In particular, fire models such as Behave~\cite{burgan1984behave,rothermel1986modeling} are capable of encapsulating fire dynamics across landscapes.} However, long-term wildfire activity prediction is fundamentally complex because of the high dimension of the data and the dynamics between wildfire activities and environmental conditions. Therefore various wildfire models have been developed at regional or global scales. The ones applied at regional scales can be used to model wildfire events in given ecoregions~\cite{ALEXANDRIDIS2008}. On the other hand, global wildfire models attempt to analyze wildfire occurrences and predict their probability density~\cite{kelley_2019_how, burton_2019_representation, mangeon_2016_inferno}. According to ~\cite{sullivan2009wildland1,sullivan2009wildland2}, \rev{wildfire models can be mainly split into two categories: physics-based and data-driven models. The latter also includes empirical models}. 

\rev{Physics-based models attempt to understand and reconstitute the dynamic relationship between wildfire activities and environmental factors through physical equations.} Physics-based models have been widely used in environmental science such as the use of wave equations to model storm runoff~\cite{costabile_2012_a}, \ac{ODE} in wind speed prediction~\cite{ye_2022_dynamicnet}, or 3D \ac{CFD} and \ac{CA} for wildfire propagation~\cite{hilton_2018_incorporating, valero2021multifidelity}. \rev{Physics-based modelling is also crucial for various climate or land surface models like JULES~\cite{sellar2019ukesm1}}, which \rev{simulates global vegetation cover, carbon and moisture exchange between the atmosphere, biosphere, and soil, and} can predict the burnt area and fire emissions at a global scale depending \rev{on environmental variables~\cite{mangeon_2016_inferno, burton_2019_representation}. In addition, hybrid coupled-atmosphere wildfire models like WRF-SFIRE~\cite{mandel2011coupled} and CAWFE~\cite{clark2004description} enhance prediction accuracy and computational efficiency by combining physical modelling with dynamic atmospheric data integration, often outperforming fully physical models}. However, those models \rev{typically also} rely highly on \rev{empirical} parameterization\rev{s of unresolved processes} to reach accurate results. \rev{Although some physics-based models} \rev{show promising prediction results~\cite{li2013quantifying,knorr2016climate,melton2016competition}}, the computational burden to solve those equations has made them mainly regional-specific, making these models impractical for \rev{rapid decision-making}~\cite{zhu_2014_parameterization, teckentrup_2018_simulations}\rev{, for instance to explore many different future climate or policy scenarios}. \rev{In addition, ensemble predictions~\cite{rosadi2021prediction, yoo2023rapid}, sensitivity analysis~\cite{valero2021multifidelity} and parameter calibration are~\cite{leung2020calibrating} often desired for wildfire and climate models. \rev{These tasks often require a large number of evaluations of the forward model, making such simulation extremely computationally costly, if not infeasible.}}

On the other hand, data-driven models try to best mimic physics-based models' behaviour by learning statistical representations~\cite{fu2023data,lucor2022simple}. Given the same inputs, those models \rev{might} learn through regression and \ac{ML} techniques, how physics-based models link driving variables such as environmental conditions and wildfire activity~\cite{radke_2019_firecast,  mandel2008wildland,mandel2019interactive,filippi2011simulation,cheng2022parameter}. Improvements in remote-sensing technologies, numerical weather prediction and climate models enhance the performance of data-driven models, which rely heavily on the quality and quantity of available data. As a consequence, they offer access to a large panel of various data with finer resolutions and longer forecasting~\cite{bauer_2015_the,vilar2021modelling,masoudvaziri2021streamlined}. Consequently various \ac{ML} techniques are now used in environmental science such as \ac{ANN} and \ac{SVM} for tornadoes prediction and detection~\cite{marzban_1996_a, trafalis_2003_tornado,farguell2021machine}, \ac{RF} for severe weather forecasting~\cite{hill_2020_forecasting,rohmer2018casting}, or \ac{RNN} surrogate models for predicting wildfires~\cite{kc_2021_a, cheng_2022_datadriven, zhu_2022_building}, storms~\cite{kim_2014_a, jia_2013_kriging} and floods~\cite{bermdez_2018_development, bass_2018_surrogate} activities. Nevertheless, the computational cost for large dynamic system prediction can sometimes remain heavy. Thus, recently, it is common to apply \ac{ML} approaches relying on top of \ac{ROM} techniques such as \ac{PCA}~\cite{saghri_2011_early,gong2022data}, orthogonal decomposition~\cite{argaud2018sensor,xiao2016non, xiao2014non,fu2023physics}, entropy-based compression~\cite{cheng2021observation} or \ac{ML} methods like \acp{AE}~\cite{vinuesa2022enhancing,cheng2023machine}. These methods try to summarise \rev{high-dimensional} arrays to a few principal latent features while keeping a high accuracy of reconstruction.
However, most of these \rev{data-driven} models mimic a regional-specific numerical model. To the best knowledge of the authors, none of these \ac{ML} surrogate modelling has yet been applied to surrogate global wildfire prediction models and study wildfire occurrence probability at a global scale.

In this study, we propose temporal-spatial surrogate models for JULES-INFERNO burnt area to enable fast wildfire forecasting on a global scale. These models used monthly collected/simulated data of soil moisture, vegetation, temperature and \rev{previous area burnt as
input to predict the subsequent} fire burned area on an iterative basis. Different simulations issued from a variety of initial conditions are split into a training and a test dataset. \rev{Our objective is to develop a highly efficient surrogate model that accelerates the online prediction process for the JULES-INFERNO system. To achieve this, we employ varied sets of initial conditions during the training and testing phases, enabling robust performance across different scenarios.}
%Two deep learning workflows, based on CAE-LSTM and ConvLSTM, respectively, are proposed in this work.
This work proposes two deep leaning models to train the surrogate model of JULES-INFERNO. One is based on \ac{CAE} and \ac{LSTM} (named CAE-LSTM) and another is based on convolutional \ac{LSTM} (named ConvLSTM).

%More precisely, for the former approach, by training a \ac{CAE}~\cite{ghasedi2017deep,cheng_2022_datadriven}, we first compress the \rev{high-dimensional} data into low-dimensional latent spaces while keeping maximum physical information. The quality of data compression can be evaluated by computing the reconstruction error after decompressing latent vectors back to the full physical space. A predictive model based on \ac{RNN}, more precisely the \ac{LSTM} network~\cite{hochreiter1997long}, is then applied to forecast subsequent burned area in the latent space. As for the ConvLSTM approach~\cite{shi2015convolutional}, it is a joint model which consists of both \acp{CNN} and \acp{RNN} in a single network structure, by integrating convolutional operations in the \ac{LSTM} cells. This model takes directly the spatially distributed variables as inputs and predict the global burned area as output. Thanks to the convolutional operators which manage to capture spatial patterns, it is shown in a variety of applications~\cite{shi2015convolutional,agga2021short} that ConvLSTM can outperform fully connected \ac{LSTM} networks.

To enhance the performance of the proposed models on unseen scenarios with a different range of initial parameters in the test dataset, fine tuning strategies are also proposed in this work. The idea is to fine tune the developed models using simulation data (for example, $10\%$) from the beginning of the unseen scenarios to improve the future predictions. 
Numerical results in this work demonstrate that both proposed models achieve a good approximation of the JULES-INFERNO model of burnt area prediction at a global scale with a Average Error per Pixel (AEP) under 0.3\% and a Structural Similarity
Index Measure (SSIM) over 98\% compared to the outputs of JULES-INFERNO. More importantly, for both approaches, it takes roughly 10 seconds to predict the bunred area of 30 years on a laptop CPU. In contrast, running JULES-INFERNO software will cost about five hours of computational time on 32 threads with the \emph{JASMIN national High Performance Computing (HPC) system}~\cite{lawrence2013storing}.

The paper is organized as follows. The generation and the pre-processing of the training and test dataset using JULES-INFERNO are described in Section \ref{PRESENTATION OF THE DATA}. We then introduce the methodology used for computing and fine tuning the two surrogate models in Section  \ref{METHODOLOGY}. The numerical results of predicting unseen scenarios with different initial conditions are shown and discussed in Section  \ref{RESULTS}.  
Finally, we close the paper with a conclusion in Section  \ref{conlusion}.

\section{Model and dataset} \label{PRESENTATION OF THE DATA}

\indent In this section, we present the data used for training and testing our temporal-spatial surrogate models, which are generated using the JULES-INFERNO model. 

\subsection{\emph{JULES - INFERNO} model} \label{JULES-INFERNO}

\indent JULES-INFERNO is a computational vegetation and wildfire model combining the fire parameterisation INFERNO and the land surface model JULES. In JULES-INFERNO, JULES \rev{vegetation and land surface} outputs are considered as input variables of INFERNO to forecast wildfire occurrence and emissions at a global scale~\cite{mangeon_2016_inferno}.
More precisely, \ac{INFERNO} follows the simplified parameterisation for fire counts, suggested by~\cite{pechony_2009_fire}, which models fire occurrence as a relationship between fuel flammability and ignitions. Fuel flammability is a function of temperature, precipitation\rev{,} and \ac{RH}. Fire ignitions are anthropogenic (human population density) or natural (lightning).
To simulate global area burnt and emissions, \ac{INFERNO} adds \rev{additional} inputs in the flammability parameterisation scheme such as the first layer of soil moisture\rev{,} and fuel load represented by vegetation carbon density~\cite{mangeon_2016_inferno}. Average burnt area per fire is then modelled as a function of vegetation type, since  wildfires tend to be larger, for example, in grasslands than in forests~\cite{chuvieco_2008_global, giglio_2013_analysis}. \rev{The JULES-INFERNO model and it's underlying parameterizations themselves have previously been described in depth and validated with respect to global burnt area observations and other global fire models~\cite{pechony_2009_fire, mangeon_2016_inferno, burton_2019_representation, sellar2019ukesm1, teckentrup2019response, hantson2020quantitative}. Experiments have demonstrated that JULES-INFERNO performs effectively in replicating observed global burnt areas and exhibits comparable performance when compared to other widely-used global fire models.}

The \ac{JULES} model simulates on a global scale the state of land surface and soil hydrology. It considers vegetation dynamics, carbon cycle as well as the exchange of the fluxes between vegetation and environment~\cite{burton_2019_representation,best2011joint,clark2011joint2}.
These fields are therefore used as JULES-INFERNO's topsoil moisture and fuel load inputs. \ac{JULES}  uses a \ac{DGVM} called \ac{TRIFFID} to predict changes in biomass and fractional coverage of 13 different plant functional types (PFTs).

 The underlying equations of the INFERNO scheme are detailed in Section $2.1$ of ~\cite{mangeon_2016_inferno} \rev{and in~\cite{burton_2019_representation}}. With this approach, JULES-INFERNO is \rev{effective} in capturing global burnt area and diagnosing wildfire occurrences~\cite{mangeon_2016_inferno,teckentrup2019response,hantson2020quantitative}.
\rev{The JULES-INFERNO fire simulation model could be time-consuming for high-resolution or long-term ensemble simulations due to its use of complex computational algorithms, requiring iteratively solving a large set of coupled equations in order to simulate the evolution of the global land surface and biosphere, and consequent fire behaviour. Additionally, the need to simulate fire behaviour over extended periods under future climate scenarios further increases the computational time.}
\subsection{Data generation} \label{DATA GENERATION}

The objective of this study is to build efficient surrogate models for the long-term prediction of global area burnt. Four spatially distributed environmental variables of JULES-INFERNO are considered in this study : 
\begin{itemize}
\item $\bf{X}$ : field of \emph{Total area burnt} in fraction of grid-box $s^{-1}$ 
\item $\bf{V}$ : field of \emph{LAI} (Leaf Area Index) - a unitless vegetation density indicator
\item $\bf{M}$ : field of \emph{Soil moisture} in $kg m^{-2}$
\item $\bf{T}$ : field of \emph{Surface air temperature} in $K$
\end{itemize}

\rev{As the aim of our approach is computational efficiency, we choose V, M, and T as a minimal set of predictor variables which represent the leading-order controls on wildfire burnt area of fuel availability and dryness~\cite{haas2022global}.  M and T are both used explicitly as predictors of vegetation flammability in INFERNO, and are also closely related to the additional meteorological variables of relative humidity and precipitation rate. 
 V is closely related to the leaf and litter carbon pools which are used by INFERNO, but using LAI in our surrogate model allows the resulting model to be easily generalised to work with data from other DGVMs or remote sensing. } In this study, \rev{the output resolution of the JULES-INFERNO model is fixed as $112  \times 192 $ on the global map where 112 is the number of pixels on the latitude axis and 192 is the number of pixels on the longitude axis.} 
To train and test the surrogate models, \rev{we use output from} five \rev{30-year} simulations ($P_{1}$, $P_{2}$, $P_{3}$, $P_{4}$, $P_{5}$) of JULES-INFERNO\rev{.  These different simulations were each} performed with different initial conditions as summarized in Table~\ref{tab:IC_table}. \rev{For each simulation of 30 years, data are saved monthly, resulting in total 360 snapshots for each variable in each of the five simulations.}

\begin{table*}[h!]
\caption{Initial conditions for \emph{JULES-INFERNO} simulations}
\centering
\resizebox{\textwidth}{!}{
\begin{tabular}{M{2cm}M{6cm}M{6cm}M{3cm}}
\toprule
Simulation & Meteorology & Initial conditions & Ignitions \\
\midrule
$P_{1}$ & FireMIP last glacial maximum (LGM): detrended $1961$-$1990$  - NCEP reanalysis, merged with monthly LGM $20^{th}$ Century climate anomalies from MIROC PMIP-$3$ archive~\cite{rabin2017fire}  & $1200$-year spin-up repeating the $1961$-$1990$ timeseries originally initialised from an arbitrary present-day JULES run & Natural (lightning)\\

\hline

$P_{2}$ & same as $P_{1}$ & Jan $1^{st} \ 1991$ from $P_{1}$ & Natural (lightning) \\

\hline

$P_{3}$ & same as $P_{1}$  & Jan $1^{st} \ 1991$ from $P_{2}$ & Natural (lightning) \\

\hline

$P_{4}$ & same as $P_{1}$ & Jan $1^{st} \ 1991$ from $P_{3}$ & Natural (lightning) \\

\hline

$P_{5}$ & $1990$-$2019$ JRA reanalysis~\cite{teckentrup2021assessing} & Continuation of $1700$-$2020$ historical simulation - Original $1700$ initial conditions would have been following a $1000$-year spin-up with repeated $1700$ conditions & Natural (lightning) + anthropogenic (function of population density) \\
\bottomrule
\end{tabular}
}
\label{tab:IC_table}
\end{table*}

$P_{1}$ to $P_{4}$ simulate \rev{a nominal time} period from $1961$ to $1990$\rev{, however with shifted and detrended meteorological boundary conditions that represent a cooler climate state, taken from the FireMIP last glacial maximum (LGM) experiment~\cite{rabin2017fire}.  Mean}while $P_{5}$ corresponds to the \rev{historical} period of $1990$ to $2019$\rev{, and is taken from an experiment run under the TRENDYv9 protocol for the Global Carbon Budget 2020 report~\cite{global_c_budget_2020}}. The simulation snapshots are denoted as:

\begin{align}
    \rev{\bf{X}^{Ps}_{t} = \{\bf{X}^{Ps}_{t}\} \quad \textrm{with} \quad t\in \{ 1,...,360\} \ \textrm{and} \quad s\in \{ 1,...,5\}}.
\end{align}
\rev{Same definitions are made for $\bf{X}^{Ps}_{t}, \bf{V}^{Ps}_{t}, \bf{M}^{Ps}_{t}$ and $\bf{T}^{Ps}_{t}$.}

\subsection{Data preprocessing} \label{DATA PROCESSING}

\indent The five simulations with four variables and in total 9000 snapshots are split into a training set, a validation set and a test set. 
More precisely, the $3$ first simulations $P_{1}$, $P_{2}$ and $P_{3}$ are used to train the models, i.e.,
\begin{align}
\rev{\bf{X}^{train} = \{\bf{X}^{P_{1}}_{t}\} \ \cup \ \ \{\bf{X}^{P_{2}}_{t}\} \ \cup \ \ \{\bf{X}^{P_{3}}_{t}\}},
\end{align}
\rev{with similar definitions for $\bf{V}^{train}, \bf{M}^{train}, \bf{T}^{train}$}.

Then $P_{4}$ is used to validate the models and select the most appropriate hyperparameters. Finally, $P_{5}$, with significantly different initial conditions, is used for fine-tuning and testing the surrogate model performance, i.e., 

\begin{align}
\rev{\bf{X}^{val} = \{\bf{X}^{P_{4}}_{t}\} \quad \textrm{and} \quad \bf{X}^{FT} = \{\bf{X}^{P_{5}}_{t}\}.}
\end{align}
\rev{with similar definitions for $\bf{V}^{val}, \bf{M}^{val}, \bf{T}^{val}$ and $\bf{V}^{FT}, \bf{M}^{FT}, \bf{T}^{FT}$ }.

\begin{figure}[h!]
  \centering
  \includegraphics[width=0.5\textwidth]{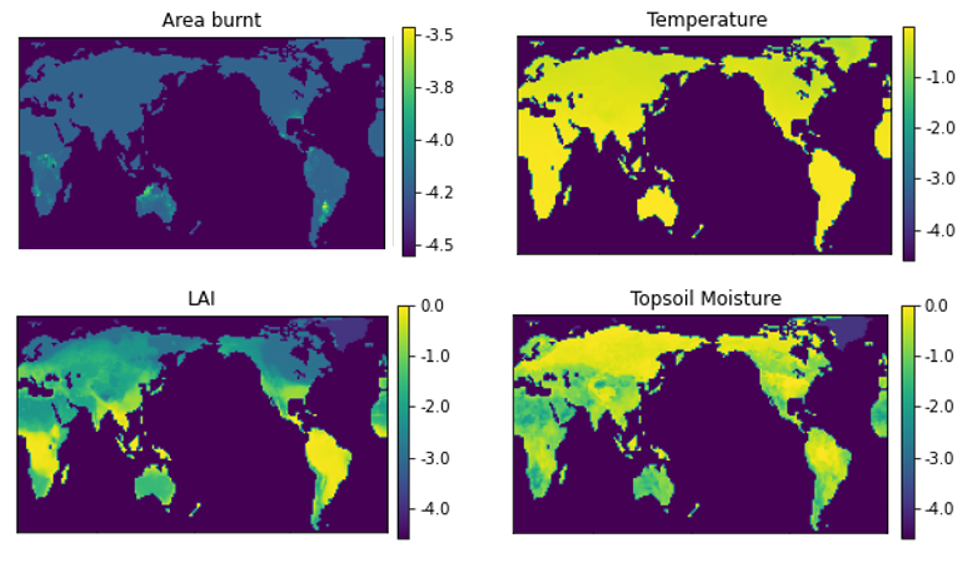}
\caption{Four spatially distributed variables in Logarithmic scale at $t = 300$ }
\label{fig:var_plot}
\end{figure}

During the training process, all four variables are normalized to the range of 0 to 1 so that they can be equally weighted in the training loss. For example, the normalization of the  \emph{Total Area burnt} leads to

\begin{equation}
    \label{eq:MinMax-scaling}
    \bf{\widehat{X}}_{t} = \frac{\bf{X}_{t} - \bf{X}^{train}_{min}}{\bf{X}^{train}_{max} - \bf{X}^{train}_{min}}
\end{equation}

where $\bf{\widehat{X}}_{t}$ is the normalized \emph{Total Area burnt}. An example of the normalized spatially distributed variables is displayed in Fig.~\ref{fig:var_plot} with a \emph{Logarithmic scale}. A land mask is applied to highlight inland points.

\begin{comment}

Finally, training sets are reorganised to associate each training sample with its output, corresponding respectively to a series of $p$ previous and $n$ subsequent forecasts. For example, the training set of $1^{st}$ simulation \emph{Total Area burnt} is :

\begin{itemize}
    \item $\bf{\dot{X}^{P_{1}}_{t}} = \{\bf{X}^{P_{1}}_{t - p}, ..., \bf{X}^{P_{1}}_{t - 1}, \bf{X}^{P_{1}}_{t}\} \ with \ \bf{t}\in \{p, ..., 360 - n\}$
    \item $\bf{\dot{X}^{YP_{1}}_{t}} = \{\bf{X}^{YP_{1}}_{t + 1}, \bf{X}^{YP_{1}}_{t + 2}, ...,  \bf{X}^{YP_{1}}_{t + n}\} \ with \ \bf{t}\in \{p+1, ..., 360 - n\}$
\end{itemize}

and then,

\begin{itemize}
    \item $\bf{\dot{X}}^{train} = \{\bf{\dot{X}}^{P_{1}}_{t}\} \ \cup \ \ \{\bf{\dot{X}}^{P_{2}}_{t}\} \ \cup \ \ \{\bf{\dot{X}}^{P_{3}}_{t}\}$
    \item $\bf{\dot{X}}^{target} = \{\bf{\dot{X}}^{YP_{1}}_{t}\} \ \cup \ \ \{\bf{\dot{X}}^{YP_{2}}_{t}\} \ \cup \ \ \{\bf{\dot{X}}^{YP_{3}}_{t}\}$
\end{itemize}

\end{comment}

\section{Methodology} \label{METHODOLOGY}
In this section, we describe the computation of the two surrogate models and the fine tuning strategies when applying these models to unseen scenarios. 
 Both models use a sequence-to-sequence prediction mechanism which takes $p$ previous time steps as inputs and return the prediction of the $n$ following time steps as outputs at each iteration.

\subsection{CAE - LSTM} \label{CAE - LSTM}

\begin{figure}[ht!]
  \includegraphics[width=0.5\textwidth]{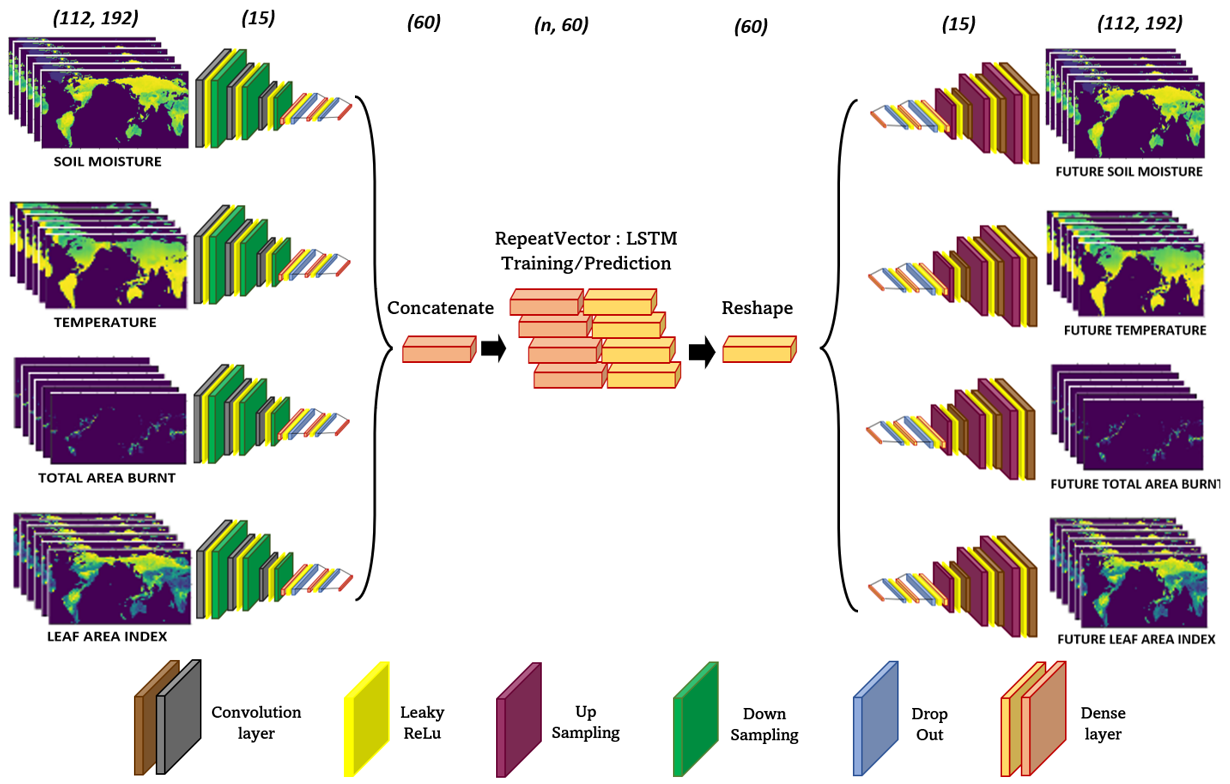}
  \centering
\caption{Workflow of Joint CAE-LSTM}
\label{fig:CAE_LSTM_arch}
\end{figure}

The CAE-LSTM applies Convolutional Autoencoder and Long-Short-Term-memory networks to 
reduce the dimension of the data and perform predictions in the reduced latent space successively. 
 Fig~\ref{fig:CAE_LSTM_arch} presents the workflow of this method with four environmental variables including the global burnt area. 

\subsubsection{CAE} \label{SECTION_CAE}

\indent The first part of the CAE-LSTM model is the \ac{CAE}, used to compress the full space data into a reduced latent space with a minimum loss of information. \ac{CAE} is a self-supervised approach based on \acp{CNN} to capture the spatial patterns. The dimension of the latent space is fixed as 15 in this study, yielding a compression rate of $\frac{12}{112 \times 192} = 0.07\% $. \rev{The dimension of the latent space is considered as a hyperparameter in this study and it is determined by numerical experiments following an analysis of principle components.}

\ac{CAE} consists of two sub-networks: the \emph{Encoder} \(\mathcal{E}\) which compresses the input data into latent variables (of dimension 15 in this study), and the \emph{Decoder} \(\mathcal{D}\) which decompresses the latent variables back to their original form ($112 \times 192$ in this study). Architectures for the \emph{Encoder} and \emph{Decoder} are various. However, they are generally composed of convolutional, pooling and dense layers. Convolutional layers manage to extract local multi-dimensional patterns thanks to the convolutional filters. Pooling layers filter the essential features to propagate in the network~\cite{rawat_2017_deep} and either reduce the convolved tensor dimensions by sub-sampling (the case of \emph{Encoder}) or, on the contrary, increase them by up-sampling (the case of \emph{Decoder}). As the final step of the \emph{Encoder}, fully connected dense layers flatten multi-dimensional tensors into a $1D$ vector of the target dimension. 
 In this study, the \emph{Decoder} is built as the inverse of the \emph{Encoder} to reconstruct spatially distributed inputs from the compressed latent variables.
%The exact architecture of the \ac{CAE} is available in Table \ref{table: CAE_structure} in the appendix of this paper.

Relying on the same structure, 4 \acp{CAE} are trained separately for each of the $4$ environmental variables $X^{train}$, $V^{train}$, $M^{train}$ and $T^{train}$ defined in Section \ref{DATA PROCESSING}. The \emph{Encoders} and the \emph{Decoders} are trained jointly, using the \emph{Adam} optimizer (\cite{bock2019proof}) and the \emph{MSE} loss function with $20\%$ of the data assigned to a validation set. The training process continues as long as the validation loss decreases.
\begin{comment}

For example, for the \emph{Total Area burnt}:

\begin{equation}
    \bf{(X^{train}_{t=1})^{C}} = \mathcal{E}(\bf{X^{train}_{t=1}}) \hspace{0.45cm} \ and \  \bf{(X^{train}_{t=1})^{Dc}} = \mathcal{E}(\bf{(X^{train}_{t=1})^{C}})
\end{equation}
where $\bf{X}^{train}_{t=1}$ and $\bf{(X^{train}_{t=1})^{Dc}}$ are of dimension $(1, \ 112, \ 192)$, and $\bf{(X^{train}_{t=1})^C}$'s is $(1, \ 15)$.
\end{comment}

The performance of data compression methods will be evaluated on unseen scenarios using \ac{SSIM} and \ac{AEP}  as presented in \textsc{Section} \ref{RESULTS}.
\subsubsection{LSTM} \label{SECTION_LSTM}

\indent  Once data compression is achieved, as the second stage of CAE-LSTM, \ac{LSTM} is used to forecast the dynamics of the latent variables. As a variant of \ac{RNN}, \ac{LSTM} has been widely  applied in the prediction of time series data or dynamical systems~\cite{mohan2018deep}. In particular, compared to standard \acp{RNN}, \ac{LSTM} uses a selective memory, making them perfectly suited to harness data with long-term dependencies~\cite{mohan2018deep}.
Furthermore, thanks to the gate structure, \ac{LSTM} can handle the vanishing gradient problem~\cite{hochreiter1998vanishing} which is cumbersome for traditional \acp{RNN}.
More precisely, $3$ types of \emph{gates} are used: the \emph{input}, the \emph{output} and the \emph{forget gates}. 
We denote $x_{t}$ and $y_{t}$ the input and the output of a \ac{LSTM} cell at time step t.
Each \ac{LSTM} cell adopts $\bf{x}_{t}$ and $\bf{y}_{t-1}$ through the \emph{input gate} $\bf{i}_t$. The cell state $\bf{c}_t$, the \emph{forget gate} $\bf{f}_t$ and the \emph{output gate} $\bf{o}_t$ are updated accordingly,
\begin{gather}
\begin{gathered}
    \label{eq:LSTM_fcts}
    {\bf{f}_{t}} = \sigma (\bf{W}_{f} \cdot [\bf{y}_{t-1}, \bf{x}_{t}] + b_{f}), \\
    {\bf{i}_{t}} = \sigma (\bf{W}_{i} \cdot [\bf{y}_{t-1}, \bf{x}_{t}] + b_{i}), \\
    {\bf{o}_{t}} = \sigma (\bf{W}_{o} \cdot [\bf{y}_{t-1}, \bf{x}_{t}] + b_{o}), \\
    {\bf{\tilde{c}}_{t}} = tanh \ (\bf{W}_{c} \cdot [\bf{h}_{t-1}, \bf{x}_{t}] + b_{c}), \\
    {\bf{c}_{t}} = \bf{f}_{t} * \bf{c}_{t-1} + \bf{i}_{t} * \bf{\tilde{c}}_{t},
\end{gathered}
\end{gather}

where $*$ denotes a matrix multiplication. $(\bf{W}_f,b_f), (\bf{W}_i,b_i), (\bf{W}_o,b_o), (\bf{W}_c,b_c)$ are the weights and the bias for each gate, respectively, updated by back-propagation during the training process. $\bf{\tilde{c}}_t$ is the updated cell state propagated through the network. The output $\bf{y}_{t}$ is then computed as
\begin{gather}
    \label{eq:LSTM_output}
    {\bf{y}_{t}} = {\bf{o_{t}}} \ * \ tanh \ ({\bf{c_{t}}}).
\end{gather}

In this study, for comparison, 2 \ac{CAE}-\ac{LSTM} models are implemented, i.e.,
\begin{itemize}
    \item \emph{Single} CAE-LSTM: Only the \emph{Total Area burnt} variable is considered as model inputs and outputs.
    \item \emph{Joint} CAE-LSTM: The four environmental variables are concatenated in the latent space and considered as model inputs and outputs (as shown in Fig~\ref{fig:CAE_LSTM_arch}).
\end{itemize}

 Both models are trained with input and output length $p$ and $n$ respectively set to $1$, $3$, $6$ and $12$ months. 
Similar to the training of \ac{CAE}, the validation set takes $20\%$ data from the training set. The \emph{Adam} optimizer and the \emph{MSE} loss function are employed, and the models are trained as long as validation loss decreases. %The exact neural network structures of  \emph{Single} and \emph{Joint} CAE-LSTM are illustrated in Table \ref{tab:LSTM Architecture} in the appendix.

\subsection{ConvLSTM} \label{ConvLSTM}

The CAE-LSTM structure is widely applied in surrogate \rev{modelling}~\cite{tang2020deep, cheng_2022_datadriven}. However, the implementation of data compression and dynamics forecasting through two separate networks increases over-fitting risk and complicates the fine-tuning process. Therefore, the second surrogate \rev{modelling} in this study use ConvLSTM networks~\cite{shi2015convolutional} which combines \ac{CNN} and \ac{LSTM} in a single network structure.  

Similar to \ac{LSTM} models, ConvLSTM uses selective memory to capture temporal-spatial patterns from multi-dimensional inputs. The strength of this model has been widely demonstrated in harnessing multi-dimensional data with temporal-spatial dependencies, such as video prediction~\cite{zhou2020deep}, image recognition~\cite{zhang2018attention} and $3D$ ocean temperature prediction~\cite{zhang2020prediction}. \rev{It has also been applied to wildfire prediction in previous studies~\cite{huot2020deep,kondylatos2022wildfire}}.

 The matrix multiplication used in \ac{LSTM} to update the cell states and outputs is replaced by convolution operations to operate $2D$ inputs, that is,

\begin{gather}
\begin{gathered}
    \label{eq:ConvLSTM_fcts}
    {\bf{f}_{t}} = \sigma (\bf{W}_{f} \cdot [\bf{H}_{t-1}, \mathcal{X}_{t}] + b_{f}), \\
    {\bf{i}_{t}} = \sigma (\bf{W}_{i} \cdot [\bf{H}_{t-1}, \mathcal{X}_{t}] + b_{i}), \\
    {\bf{o}_{t}} = \sigma (\bf{W}_{o} \cdot [\bf{H}_{t-1}, \mathcal{X}_{t}] + b_{o}), \\
    {\bf{\tilde{C}}_{t}} = tanh \ (\bf{W}_{C} \cdot [\bf{H}_{t-1},\mathcal{X}_{t}] + b_{C}), \\
    {\bf{C}_{t}} = \bf{f}_{t} \circ \bf{C}_{t-1} + \bf{i}_{t} \circ \bf{\tilde{C}}_{t}, \\
\end{gathered}
\end{gather}

where $\circ$ denotes a convolutional operator. The output of the previous cell $\bf{H}_{t-1}$, the current input $\mathcal{X}_{t}$, the previous and new cell states, $\bf{C}_{t-1}$ and $\bf{C}_{t}$, are $2D$ tensors in ConvLSTM. The output $\bf{H}_{t}$ of the current cell, also in a $2D$ form, is computed as

\begin{gather}
\begin{gathered}
    {\bf{H}_{t}} = {\bf{o}_{t}} \ \circ \ tanh \ ({\bf{C}_{t}}).
\end{gathered}
\end{gather}

\begin{figure}[ht!]
  \centering
  \includegraphics[width=0.5\textwidth]{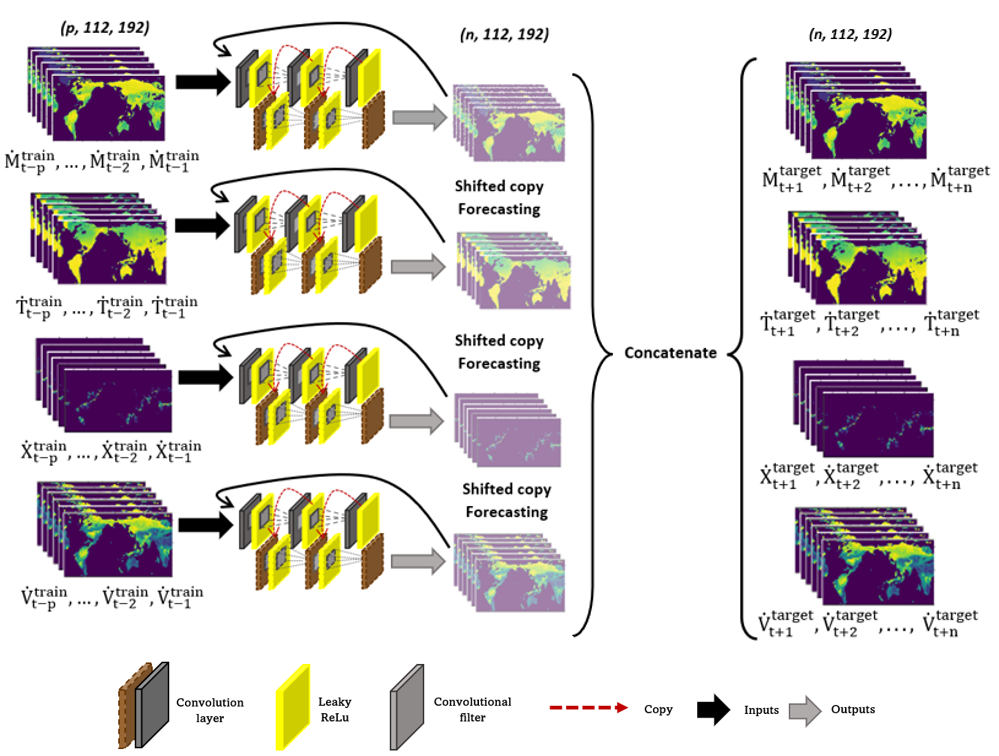}
\caption{Workflow of \emph{Joint} ConvLSTM}
\label{fig:ConvLSTM_arch}
\end{figure}
Same as CAE-LSTM,  $2$ structures of ConvLSTM are built in this study. \emph{Single} ConvLSTM which predicts only the \emph{Total Area burnt} and \emph{Joint} ConvLSTM based on the four environmental variables is shown in Fig~\ref{fig:ConvLSTM_arch}. Joint training of multi-channel temporal-spatial systems can be difficult and require large amounts of training data. Therefore, in the \emph{Joint} model, instead of considering four environmental variables as different channels, four separate ConvLSTM blocks have been implemented and concatenated before the output layer. 
For choosing the most appropriate input and output length, $n$ and $p$ are respectively set to $1$, $3$, $6$ and $12$ months. 
The validation set for both \emph{Joint} and \emph{Single} ConvLSTM training consists of $20\%$ of the training data. \emph{Adam} and \emph{MSE} are used as the optimiser and the loss function, respectively. 

\subsection{Model Fine tuning} \label{Model fine tuning}

\indent During training, \ac{ML} models optimise their parameters to best harness the training data, leading to potential risk of overfitting. Consequently those models often struggle when facing unseen data of different periods or regions and lose prediction accuracy. 

 In \ac{ML}, Fine-tuning consists of adjusting models parameters by re-training the \emph{pre-trained} models with (usually a small amount of) unseen data of different initial conditions to improve the model generalizability and robustness. In this study, the first 5 years of the test set (i.e., $P_{5}$ ($1990 - 1995$)) are used to fine-tune the pre-trained surrogate models. 
 More precisely, in our ConvLSTM model, convolutional layers and \ac{LSTM} cells are fine-tuned simultaneously thanks to the joint structure.
In our CAE-LSTM model, only the \ac{LSTM} network is fine-tuned, since fine-tuning the CAE \rev{should} require the complete re-training of the LSTM.
Since the fine-tuning dataset is of small size, to avoid overfitting,
 the number of fine-tuning epochs is fixed as $30$ in this study. Finally, fine-tuned models are tested on the last $25$ years remaining of $P_{5}$ ($1995 - 2020$). In other words, despite that the fine-tuning requires to simulate the beginning of the test sequence ($16.6\%$ in this study) using JULES-INFERNO, the online computational time can still be considerably reduced compared to running the full simulation. In this work, the training and the fine tuning of both surrogate models are performed on a Tesla P100 GPU using the Google Colab environment while the online prediction is made on a single-core CPU of 2.2GHz.  The running time of JULES-INFERNO on \textit{JASMIN national HPC} is estimated as an average from 4 simulations, range 4.3 – 5.9 hours, using Intel Xeon E5-2640-v4 "Broadwell" or Intel Xeon Gold-5118 “Skylake” processors with ~7 GB RAM available per thread.

\begin{comment}

\begin{itemize}[noitemsep]
\item $\bf{X}^{FT}$ = \{$\bf{X}^{P_{5}}_{t}$$\}_{\bf{t}\in \{1, ..., 60\}}$ \ and \ $\bf{X}^{test}$ = \{$\bf{X}^{P_{5}}_{t}$$\}_{\bf{t}\in \{61, ..., 360\}}$
\item $\bf{V}^{FT}$ = \{$\bf{V}^{P_{5}}_{t}$$\}_{\bf{t}\in \{1, ..., 60\}}$ \ and \ $\bf{V}^{test}$ = \{$\bf{V}^{P_{5}}_{t}$$\}_{\bf{t}\in \{61, ..., 360\}}$
\item $\bf{M}^{FT}$ = \{$\bf{M}^{P_{5}}_{t}$$\}_{\bf{t}\in \{1, ..., 60\}}$ \ and \ $\bf{M}^{test}$ = \{$\bf{M}^{P_{5}}_{t}$$\}_{\bf{t}\in \{61, ..., 360\}}$
\item $\bf{T}^{FT}$ = \{$\bf{T}^{P_{5}}_{t}$$\}_{\bf{t}\in \{1, ..., 60\}}$ \ and \ $\bf{T}^{test}$ = \{$\bf{T}^{P_{5}}_{t}$$\}_{\bf{t}\in \{61, ..., 360\}}$
\end{itemize}
\end{comment}

\section{Numerical results and discussions} \label{RESULTS}

\indent In this section we evaluate and discuss the performance of the surrogate models regarding a variety of different metrics. The optimal neural network structures and hyperparameters are chosen by evaluating the algorithm performance on P4. The capability of the surrogate models in predicting unseen scenarios are assessed on P5 where the boundary conditions are significantly different as shown in Table \ref{tab:IC_table}.

\subsection{Metrics} \label{METRICS}

\indent Three metrics are used in this study to measure the model performances for predicting the \emph{Total Area burnt}.  
The first metric used is the Absolute Error per Pixel (\ac{AEP}), which highlights the pixel-wise differences between original ($\bX$) and predicted ($\bX'$) fields defined as
\begin{equation}
    \label{eq:Relative error}
    AEP = \frac{\Sigma_{i=1}^{r}\Sigma_{j=1}^{q}{|\bX_{ij}-\bX'_{ij}|}}{l},
\end{equation}

where $l=7771$ is the number of land points in the image. 

However, when evaluating the \ac{AEP}, predicted and original fields are compared pixel-by-pixel which makes the estimated score highly sensitive to image distortion and translation. 

\rev{To address this limitation, the work of~\cite{wang2004image} proposed the \emph{Structural Similarity Index Measure} (\ac{SSIM}),} which measures the \emph{perceptive} similarity between images (2D  vectors). This \ac{SSIM} for two images $I_{1}$ and $I_{2}$ is defined as

\begin{equation}
    \label{eq:MSE}
    SSIM = \frac{(2\mu_{I_{1}}\mu_{I_{2}} + C_{1})(2\sigma_{I_{1}I_{2}} + C_{2})}{(\mu_{I_{1}}^2 + \mu_{I_{2}}^2 + C_{1})(\sigma_{I_{1}}^2+\sigma_{I_{2}}^2 + C_{2})},
\end{equation}

where ($\mu_{I_{1}}$, $\mu_{I_{2}}$) and ($\sigma_{I_{1}}$ ,$\sigma_{I_{2}}$) are respectively the means and the standard deviations of the two images.  $\sigma_{I_{1}I_{2}}$ is the covariance of $I_{1}$ and $I_{2}$. $C_{1}$ and $C_{2}$ are regularization constants~\cite{wang2004image}. By definition, the value of \ac{SSIM} ranges from 0 to 1 indicating the similarity between $I_{1}$ and $I_{2}$.

\begin{comment}

They are computed as :

\begin{equation}
    \label{eq:ssim_constants}
    C_1 = (k_1L)^2 \hspace{1cm} ; \hspace{1cm} C_2 = (k_2L)^2
\end{equation}

where $L$ is the dynamic range of pixel value ($255$ for \emph{$8$-bit} images), and $k_{1}$, $k_{2}$ are respectively set to $0.01$ and $0.03$ by default. \ac{SSIM} is displayed as a percentage.

However, as wildfires cannot occur on the water, a land mask sets each forecast upon water points to $-0.01$ by point-wise multiplication. AEP
is measured exclusively from land point forecasts and SSIM is evaluated after applying the land mask.

\end{comment}

Finally, the third metric is the online computational time for reconstructing or predicting \emph{Total Area burnt}.

\subsection{Evaluation of data compression}
%Before using \ac{CAE} as the first step of our \emph{CAE - LSTM} model, we verified if other \ac{ROM} methods could compress our data more accurately. %In literature, some \emph{Encoders} and \emph{Decoders} exclusively use fully connected dense layers~\cite{bank2020autoencoders}. They are called \ac{AE}. Nevertheless, \ac{AE}s imply many parameters to fit, leading to expensive computational costs for \rev{high-dimensional} data~\cite{cheng2022parameter}.

\indent  Different sets of hyperparameters of \ac{CAE} are tested and compared  to select the most appropriate network structure with minimum information loss.
\ac{PCA} and \rev{fully connected autoencoders} are also implemented in this work as baselines for comparison \rev{purposes}. 

All the data compression methods are trained on the training dataset and tested on P4. Table~\ref{tab:res_encoders} presents the \ac{SSIM}, \ac{AEP} and compression/decompression computational time for each method.
\\
\begin{table}[ht!]
    \caption{Encoders results evaluated on p4}
    \centering
    \begin{tabular}{cccc}
        \toprule
        % \multicolumn{3}{c}{Model} \\
        % \cmidrule(r){1-2}
        Encoder & time (s) & AEP& SSIM \\
        \midrule
        AE & $0.25$ & $5.92 \times 10^{-5}$ & $99.89$\\
        PCA & $0.07$ & $3.99 \times 10^{-5}$ & $99.95$ \\
        CAE & $0.59$ & $3.09 \times 10^{-5}$ & $99.97$ \\
        \bottomrule
    \end{tabular}
    \label{tab:res_encoders}
\end{table}

As displayed in Table \ref{tab:res_encoders}, the online computational time of data compression methods is less than 1 second, which can be ignored in the prediction process. In fact, the data will only need to be encoded to initialize the prediction process and decoded when full-space forecast is required. 
 While all methods show strong performance regarding \ac{SSIM} scores above $99\%$, a significant advantage of \ac{CAE} can be noticed respect to the \ac{AEP}, thanks to its capability of capturing spatial dependencies.

\subsection{Evaluation of predictive models}

\indent  Here we first compare the performance of the two surrogate models on the validation dataset in terms of forecasting the next \emph{Total Area burnt} on a global scale.
As mentioned in section \ref{METHODOLOGY}, CAE-LSTM and ConvLSTM have been both trained with solely the \emph{Total Area burnt} variable (i.e., \emph{Single} surrogate model) or the four environmental variables (i.e., \emph{Joint} surrogate model). Table \ref{tab:results_mod1} shows the mean \ac{AEP} and the mean \ac{SSIM} for different models evaluated on the 30 years of prediction on P4.

\begin{table*}[ht!]
    \centering
    \captionsetup{justification=centering}
\caption{Comparison of \emph{Single} and \emph{Joint} surrogate models on P4}
        \begin{tabular}{ccc|cc}
            \toprule
            {} & \multicolumn{2}{c}{CAE - LSTM} & \multicolumn{2}{c}{ConvLSTM} \\
            \cmidrule(r){2-5}
            Metric & \emph{Single} strategy & \emph{Joint} strategy  & \emph{Single} strategy & \emph{Joint} strategy \\
            \midrule
            mean AEP & $8.74$\ x $10^{-4}$ & $8.53$\ x $10^{-4}$ & $2.90$\ x $10^{-3}$ & $1.43$\ x $10^{-3}$ \\
            mean SSIM & $99.874\%$ & $99.895\%$ & $98.513\%$ & $99.607\%$ \\ 
            \emph{Prediction time (s)} & $17.8$ & $18.2$ & $23.4$ & $70.1$ \\
            \bottomrule
        \end{tabular}
    \label{tab:results_mod1}
\end{table*}

\begin{comment}
 
\begin{table}[ht!]
    \centering
    \captionsetup{justification=centering}
    \caption{Comparison of \emph{Joint} CAE-LSTM and \emph{Joint} ConvLSTM results following the 12to1 or 12to12 strategy}
        \begin{tabular}{ccc|cc}
            \toprule
            {} & \multicolumn{2}{c}{\emph{Joint} CAE - LSTM} & \multicolumn{2}{c}{\emph{Joint} ConvLSTM} \\
            \cmidrule(r){2-5}
            Metric & \emph{M21} & \emph{M2M} & \emph{M21} & \emph{M2M} \\
            \midrule
            AEP & $8.53$\ x $10^{-4}$ & $7.97$\ x $10^{-4}$ & $1.43$\ x $10^{-3}$ & $1.37$\ x $10^{-3}$ \\
            mean SSIM & $99.895\%$ & $99.903\%$ & $99.607\%$ & $99.659\%$ \\ 
            \emph{Prediction time (s)} & $18.2$ & $2.5$ & $70.1$ & $11$ \\
            \bottomrule
        \end{tabular}
    \label{tab:res_mod_2}
\end{table}
\end{comment}
\rev{According to the results presented in Table \ref{tab:results_mod1}, the accuracy of the Total Area burnt prediction when the four environmental variables were taken
into account.} As mentioned in~\cite{shi2022characterization,kuhn2021importance}, soil moisture, LAI and temperature can significantly impact the wildfire burned area. From a data perspective, our results numerically demonstrate this matter of fact. \rev{As shown in Table \ref{tab:results_mod1}, in particular, the \emph{Joint} model of ConvLSTM can reduce more than 50\% of AEP while keeping a low online computational time.}

\begin{figure}[h!]
    \centering
    \captionsetup{justification=centering}
    \includegraphics[width=0.5\textwidth]{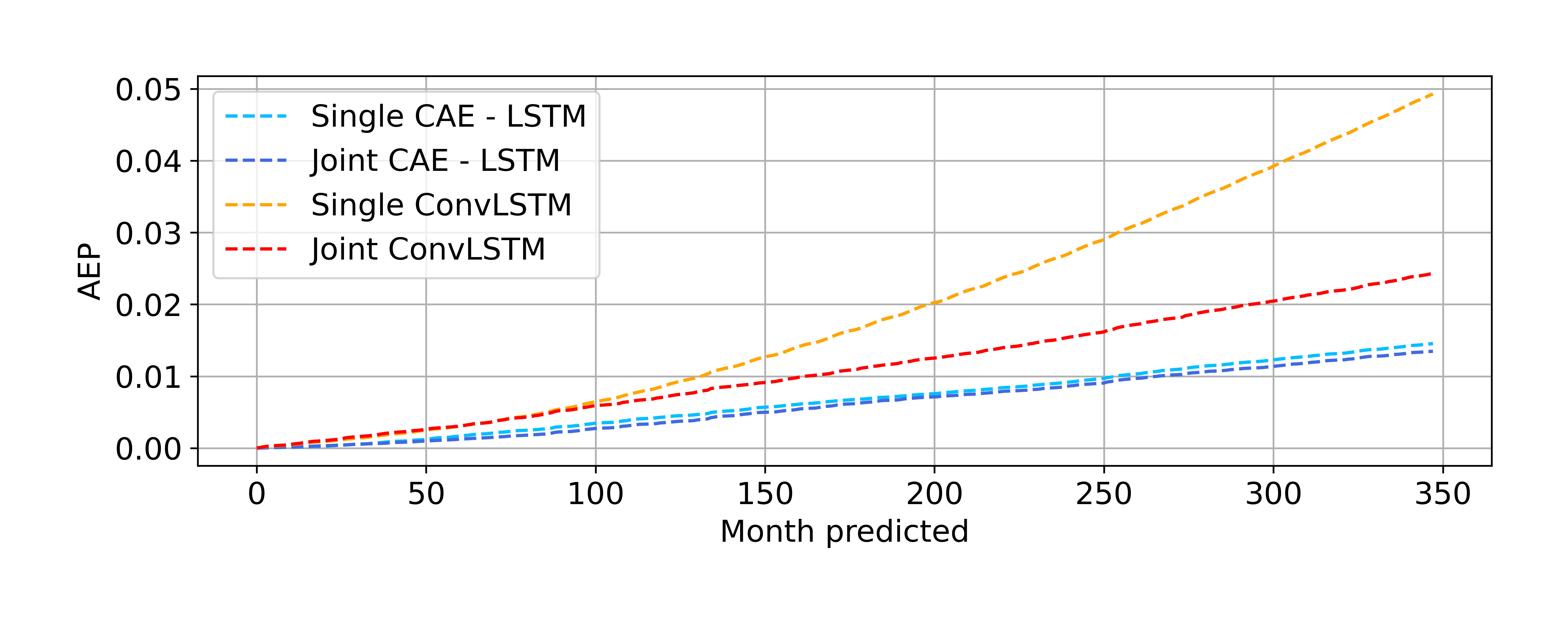}
    \caption{Cumulative sum of AEP on $P_{4}$ for \emph{CAE-LSTM} and \emph{ConvLSTM} as \emph{Joint} and \emph{Single} models}
    \label{RE_1}
\end{figure}

Fig~\ref{RE_1} presents the cumulative sum of the \ac{AEP} in CAE-LSTM and ConvLSTM predictions on the test dataset $P_{4}$ according to different models chosen. Comparing the red and the yellow dashed lines, we can conclude that including all four environmental variables in the system can significantly reduce the prediction error for ConvLSTM. On the other hand, little difference can be found between the light blue and the dark blue curves in Fig~\ref{RE_1}. This indicates that CAE-LSTM forecasting is essentially based on the previous \emph{Total Area burnt} sequences and that the contribution of other environmental variables to the predictive model is marginal, albeit still a minor improvement.
Overall, when being applied directly to unseen scenarios, CAE-LSTM shows a more robust prediction of the \emph{Total Area burnt} compared to ConvLSTM with a lower cumulative AEP.
For the rest of this paper, we will focus on the \emph{Joint} models since they are demonstrated to be more accurate compared to \emph{single} models for both CAE-LSTM and ConvLSTM.

Determining the appropriate number of monthly steps $(n,p)$ for input and output sequences is crucial for predictive models. In fact, it can be cumbersome to train predictive models with long-temporal dependencies~\cite{bengio1994learning}. On the other hand, iterative predictions using short-term forward models require frequent model forecasts, leading to more computational time, and more importantly, fast error accumulation~\cite{cheng_2022_datadriven}. Therefore, an optimal tradeoff should be found. To simplify the iterative prediction process, the input and output sequences are set to be equal (i.e.,$n=p$) in this work.

 Table~\ref{tab:res_mod_3} presents the mean \ac{SSIM} and \ac{AEP} scores for \emph{Joint} models with $n=p=3$, $n=p=6$ and $n=p=12$. It can be clearly seen that for both CAE-LSTM and ConvLSTM, the 12 to 12 setting has the best performance in terms of both prediction accuracy and computational efficiency, which is consistent with the annual periodic nature of climate variables.

\begin{table*}[ht!]
    \centering
    \captionsetup{justification=centering}
    \caption{Comparison of \emph{M2M Joint} CAE-LSTM and \emph{M2M Joint} ConvLSTM results on P4}
        \begin{tabular}{cccc|ccc}
            \toprule
            {} & \multicolumn{3}{c}{\emph{M2M Joint} CAE - LSTM} & \multicolumn{3}{c}{\emph{M2M Joint} ConvLSTM} \\
            \cmidrule(r){2-7}
            Metric & \emph{3 to 3} & \emph{6 to 6} & \emph{12 to 12} & \emph{3 to 3} & \emph{6 to 6} & \emph{12 to 12} \\
            \midrule
             mean AEP & $1.85$\ x $10^{-3}$ & $1.82$\ x $10^{-3}$ & $7.97$\ x $10^{-4}$ & $1.46$\ x $10^{-3}$ & $1.46$\ x $10^{-3}$ & $1.37$\ x $10^{-3}$ \\
            mean SSIM & $99.446\%$ & $99.408\%$ & $99.903\%$ & $99.611\%$ & $99.578\%$ & $99.659\%$ \\ 
            Prediction time (s) & $11.14$ & $7.20$ & $2.51$ & $18.10$ & $12.40$ & $11.12$ \\
            \bottomrule
        \end{tabular}
    \label{tab:res_mod_3}
\end{table*}

\begin{figure}[h!]
    \centering
    \captionsetup{justification=centering}
    \includegraphics[width=0.5\textwidth]{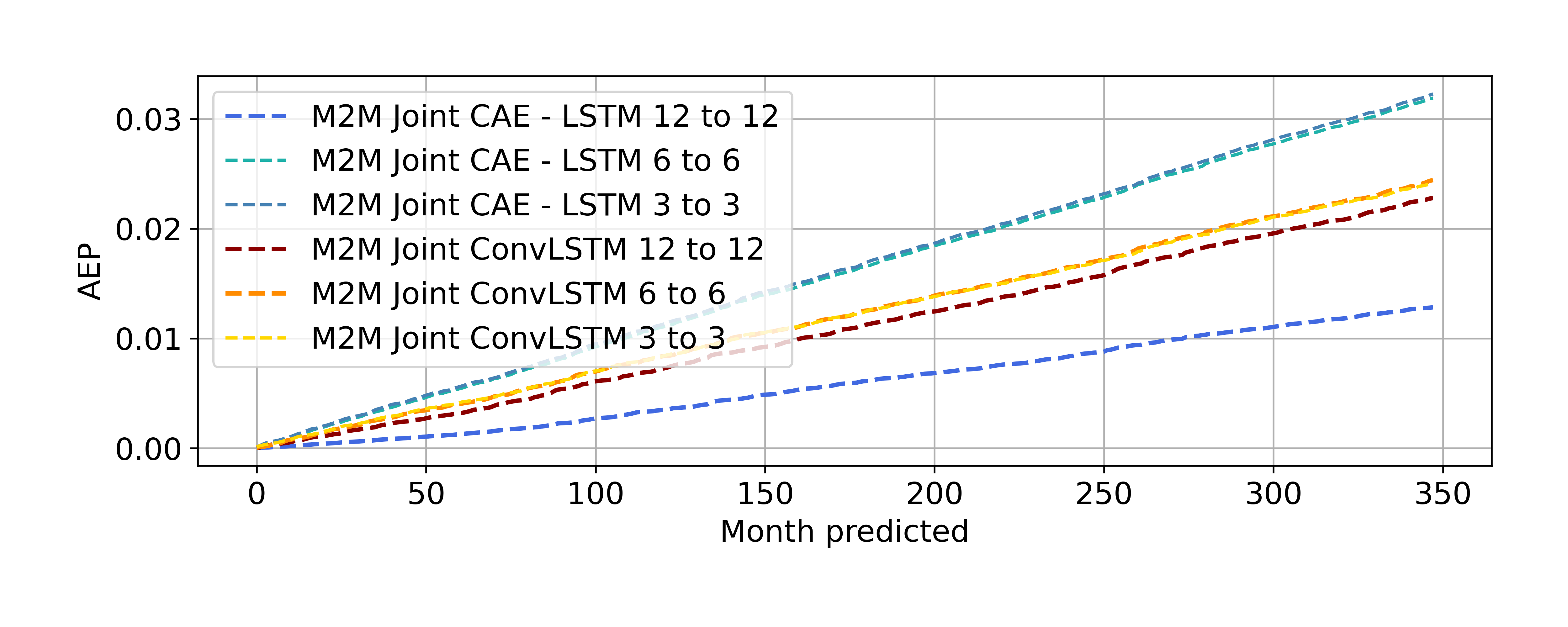}
    \caption{Cumulative sum of AEP on $P_{4}$ for \emph{M2M} \emph{Joint CAE - LSTM} and \emph{Joint ConvLSTM} with different values of $p$ and $n$}
    \label{RE_3}
\end{figure}

\rev{Similar analysis can be performed by investigating the \ac{AEP} curves as displayed in Fig. \ref{RE_3}. As can be seem there, the 12 to 12 predictive models can
lead to more reliable and consistent predictions.}
In particular, a higher sensitivity of CAE-LSTM regarding the length of input and output sequences has been noticed where both the 3 to 3 and the 6 to 6 models have an \ac{AEP} three times larger than the one of 12 to 12.

We have also tested the model performance using the data in $P_{5}$ where the simulation period and the initial conditions differ significantly from the training set as previously shown in Table \ref{tab:IC_table}.
For comparison purpose, the results of $P_5$ 
are presented alongside those of $P_4$ in Table \ref{tab:res_mod_5}. \rev{  $P_5$ corresponds to a significantly different time period and climate state from the P1-P3 simulations the algorithm was originally trained on (historical 1990-2019 versus LGM), and so is a good test of the ability of the algorithm to capture the drivers of fire under very different conditions. } Unsurprisingly, for both models, the prediction on $P_5$ is less accurate compared to $P_4$, especially in terms of \ac{AEP}. Consistent with our previous analysis, we notice that CAE-LSTM is more sensitive to the difference regarding study period and range of initial conditions. Contrary to the case of $P_4$, advantages of ConvLSTM compared to CAE-LSTM is  noticed in terms of both metrics.

\begin{figure}[ht!]
        \centering
        \includegraphics[width=0.5\textwidth]{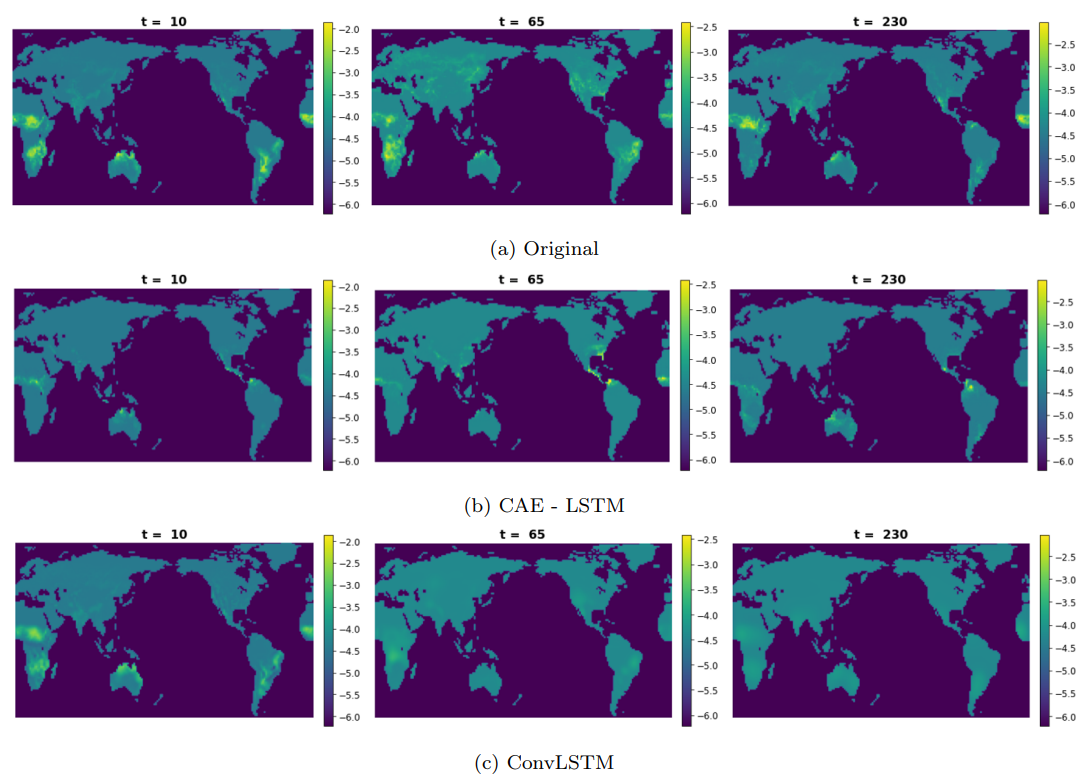}
    \caption{Surrogate models predictions of \emph{Total Area burnt} maps in $P_{5}$ for $t = 10, 65, 230$ months}
    \label{plots_pred_nft}
\end{figure}

\begin{table*}[ht!]
    \centering
    \captionsetup{justification=centering}
    \caption{Comparison of \emph{CAE - LSTM 12to12} and \emph{ConvLSTM 12to12} predictions results on $P_{4}$ and $P_{5}$}
        \begin{tabular}{ccc|cc}
            \toprule
            {} & \multicolumn{2}{c}{\emph{CAE - LSTM 12to12}} & \multicolumn{2}{c}{\emph{ ConvLSTM 12to12}} \\
            \cmidrule(r){2-5}
            Metric & $P_{4}$ & $P_{5}$ & $P_{4}$ & $P_{5}$ \\
            \midrule
            mean AEP & $7.97$\ x $10^{-4}$ & $2.13$\ x $10^{-3}$ & $1.37$\ x $10^{-3}$ & $1.73$\ x $10^{-3}$ \\
            mean SSIM & $99.90\%$ & $98.52\%$ & $99.66\%$ & $98.96\%$ \\ 
            Prediction time (s) & $2.51$ & $2.29$ & $11.12$ & $7.31$\\
            \bottomrule
        \end{tabular}
    \label{tab:res_mod_5}
\end{table*}
To further inspect the algorithm performance, \emph{Total Area burnt} has been investigated. Fig~\ref{plots_pred_nft}  shows the predicted fields of the \emph{Total Area burnt} in a logarithmic scale for t = 10, 65 and 230 months after the start of the simulation. \rev{These three time steps are chosen, because they correspond to short-, medium- and long-term prediction of burnt area, respectively.} At t = 10, ConvLSTM manages to deliver a precise prediction regarding the JULES-INFERNO output. As long as iterative predictions take place, the prediction error can be accumulated, leading to noise in future predictions. However, most at-risk regions such as \emph{Central America} and \emph{South America} at t=65, and \emph{South Africa} at t=230 can still be identified by the ConvLSTM model.  As for CAE-LSTM, the model prediction differs from the JULES-INFERNO simulation right from the beginning of the prediction process as shown in Fig~\ref{plots_pred_nft} (b). These results are coherent with the metrics shown in Table \ref{tab:res_mod_5}. In summary, despite that the CAE-LSTM surrogate model shows better performance when the test data is relatively similar (but still significantly different) to the training data in terms of time period and initial conditions (i.e., $P_4$), it is outperformed by ConvLSTM regarding the generalizability when being applied to test data with a different range of initial and meteorological conditions (i.e., $P_5$). To achieve reliable long-term predictions on $P_5$, the performance of both CAE-LSTM and ConvLSTM needs to be improved.  

\begin{comment}

\begin{figure}[ht!]
    \centering
    \captionsetup{justification=centering}
    \includegraphics[width=1\textwidth]{SSIM_4.png}
    \label{fig:cumsum1}
    \caption{Predictions SSIM scores on $P_{5}$ of CAE-LSTM 12to12 and ConvLSTM 12to12 surrogate models} 
    \label{ssim_4}
\end{figure}
\begin{figure}[h!]
    \centering
    \captionsetup{justification=centering}
    \includegraphics[width=1\textwidth]{RE_4.png}
    \caption{Cumulative sum of CAE-LSTM and ConvLSTM predictions AEPon $P_{5}$}
    \label{RE_4}
\end{figure}

\end{comment}
\subsection{Model fine tuning} \label{Calibration}

\indent Fine tuning pretrained models for unseen scenarios with significantly different conditions or assumptions is a common practice for the deployment of machine learning techniques~\cite{chu2016best,too2019comparative}. In this study, model fine-tunings are performed using the simulation data for the first five years (i.e., 60 snapshots) of $P_5$ with 30 epochs for each surrogate model.

\begin{table*}[ht!]
    \centering
    \captionsetup{justification=centering}
    \caption{Comparison of CAE-LSTM and ConvLSTM predictions on $P_{5}$ before and after fine tuning}
        \begin{tabular}{ccc|cc}
            \toprule
            {} & \multicolumn{2}{c}{CAE - LSTM 12to12} & \multicolumn{2}{c}{ConvLSTM 12to12} \\
            \cmidrule(r){2-5}
            Metric & \emph{Original} & \emph{Fine tuned} & \emph{Original} & \emph{Fine tuned} \\
            \midrule
            mean AEP & $2.13$\ x $10^{-3}$ & $2.01$\ x $10^{-3}$ & $1.73$\ x $10^{-3}$ & $1.51$\ x $10^{-3}$ \\
            mean SSIM & $98.52\%$ & $98.70\%$ & $98.96\%$ & $99.24\%$ \\ 
            Prediction time (s) & $2.29$ & $2.34$ & $7.31$ & $6.53$ \\
            \bottomrule
        \end{tabular}
    \label{tab:res_mod_4}
\end{table*}

\begin{figure}[ht!]
            \centering
            \captionsetup{justification=centering}
            \includegraphics[width=0.5\textwidth]{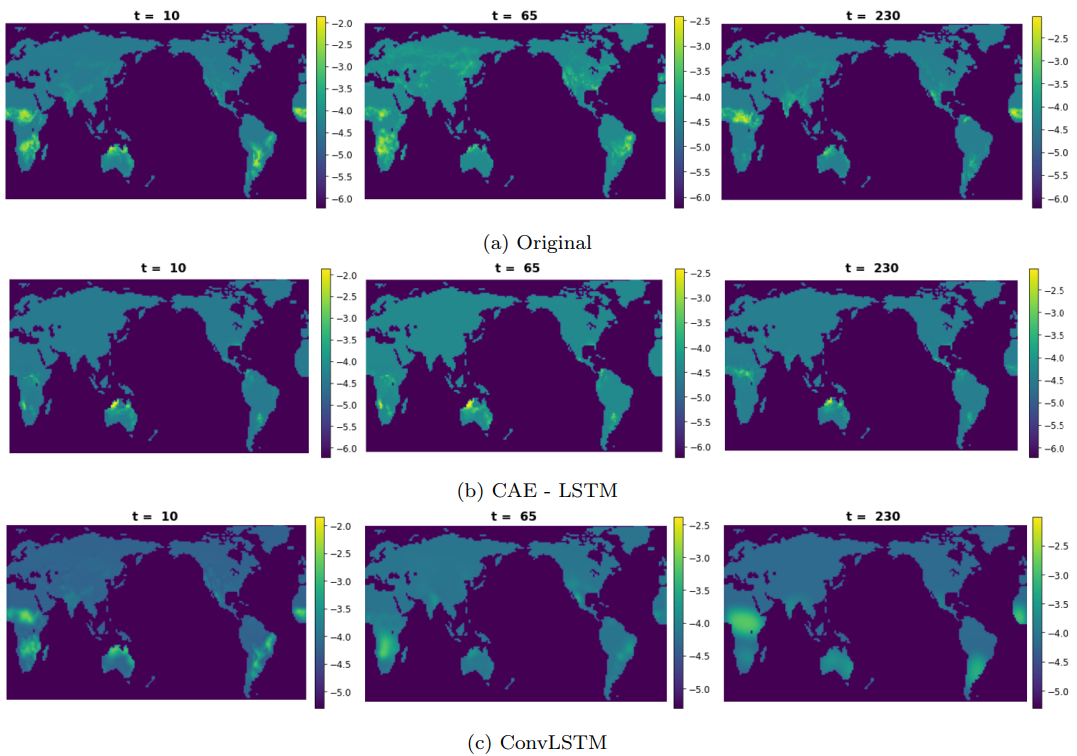}
            \label{fig:cnn_p5_ft}
    \caption{Surrogate models predictions on $P_{5}$ after fine-tuning for $t = 10, 65, 230$ months after the start of the prediction}
    \label{plots_pred_ft}
\end{figure}

\begin{figure}[ht!]
    \centering
    \captionsetup{justification=centering}
        \includegraphics[width=0.5\textwidth]{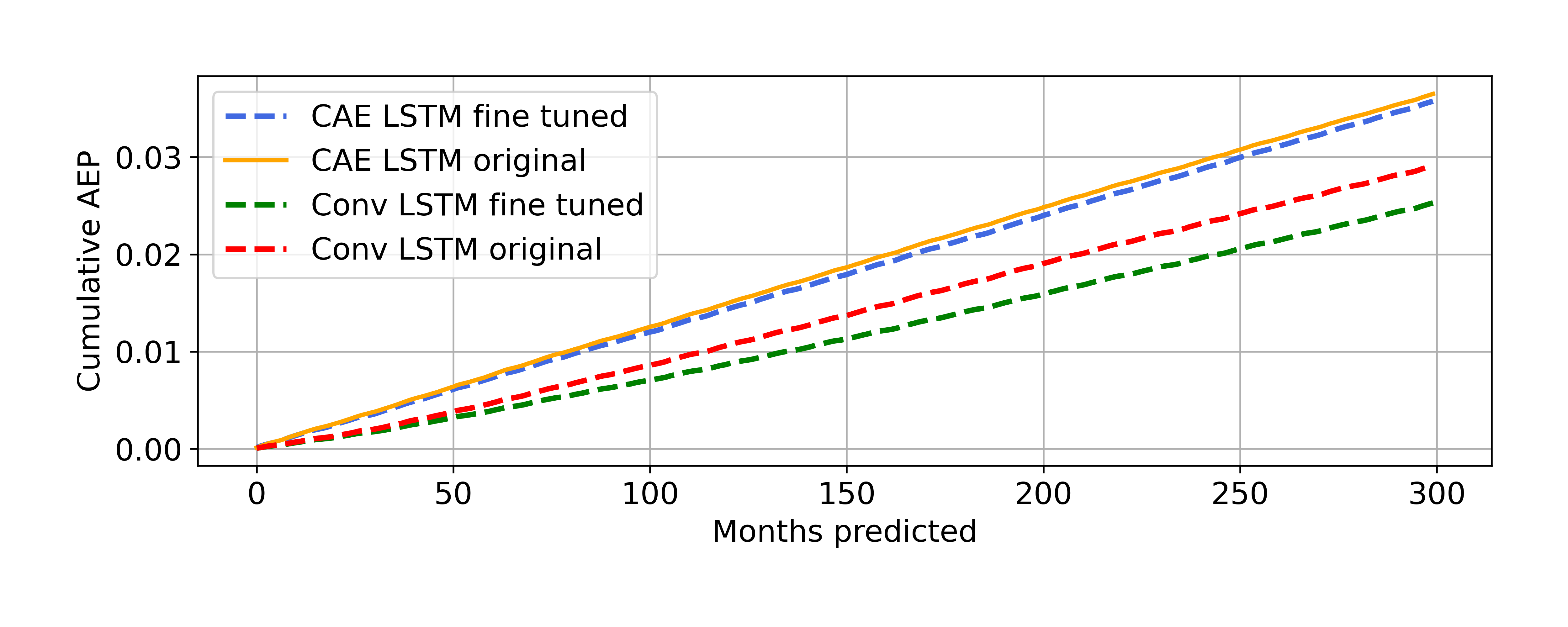}
    \includegraphics[width=0.5\textwidth]{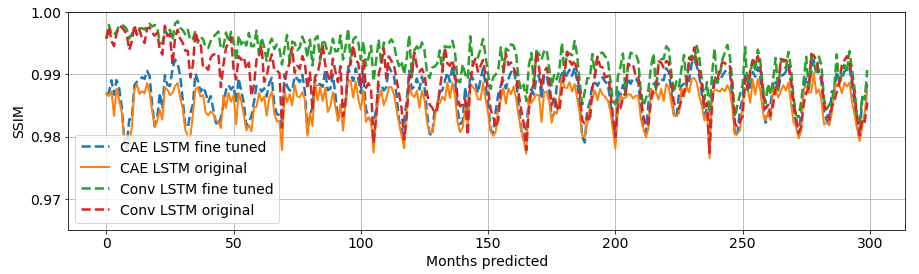}
    \caption{Cumulative AEP and SSIM of each forecast for the surrogate models before and after fine tuning}
    \label{fig:ssim_evolution}
\end{figure}

Both \ac{SSIM} and \ac{AEP} metrics are consistently improved according to the results in Table \ref{tab:res_mod_4}.
More importantly, as shown in Fig~\ref{plots_pred_ft}, considerable enhancement on long-term prediction can be noticed for both surrogate models at $t=65,230$. Most of \emph{at risk} regions (Fig~\ref{plots_pred_ft} (a)) can be successfully recognized by CAE-LSTM and ConvLSTM.  
The evolution of Cumulative AEP and \ac{SSIM} against prediction steps is shown in Fig~\ref{fig:ssim_evolution}. A consistent improvement of the \ac{SSIM} score (dashed blue line vs. solid orange line for CAE-LSTM and dashed green line vs. dashed red line for ConvLSTM) thanks to the fine tuning can be observed for both models. On the other hand, it is also noticed in Fig~\ref{plots_pred_ft} and \ref{fig:ssim_evolution} that after fine-tuning, the two stage surrogate model CAE-LSTM is still outperformed by ConvLSTM.  
In fact, the fine-tuning of ConvLSTM involves the entire neural network architecture, whereas in CAE-LSTM only the \ac{LSTM} stage is fine-tuned. Thus, the \ac{CAE} remains driven by the temporal-spatial patterns specific to the $1960 - 1990$ period and thus struggles to encode and decode data from other periods. In summary, performing fine-tuning can substantially enhance the prediction performance but also increase the computational cost since it requires running the full JULES-INFERNO model for 5 years of initial prediction. However, compared to a complete simulation of 30 years, it can still reduce the computational time from about five hours to less than an hour.

\section{Conclusion and further work}
\label{conlusion}

\indent This study presents two surrogate models of JULES-INFERNO \rev{which use} \ac{ROM} and \ac{ML} methods to speed up global burnt area forecasting. Both models, referred as CAE-LSTM and ConvLSTM, forecast monthly global area burnt from antecedent \emph{Temperature}, \emph{Vegetation}, \emph{Soil Moisture} and \emph{Total Area burnt} fields. Models are trained with JULES-INFERNO simulated forecasts from $1960$ to $1990$ but can be applied to a different period or conditions using newly available simulation or observation data. 
The numerical results in this paper demonstrate the efficiency and the robustness of the proposed approach.  Their predictions over the period of $1960$ to $1990$ (where the model has been trained) show more than $99\%$ of similarity with JULES-INFERNO simulation. The ConvLSTM-based surrogate model also shows a good generalizability after fine-tuning, thanks to its joint structure of convolutional and recurrent layers. More importantly, CAE-LSTM and ConvLSTM have considerably improved the computational efficiency. While running a 30-year simulation with JULES-INFERNO requires approximately five hours on \emph{JASMIN national HPC} ($32$ threads)  our models require less than 20 seconds on a single-core CPU. 
This paper focuses on the development of a rapid surrogate model for JULES-INFERNO, driven by the motivation to enhance efficiency. It is important to note that since the surrogate model is \rev{trained solely with} generated data from JULES-INFERNO, it will not surpass the original model in terms of accuracy. It is also important to note that the methods used in this paper can be easily extended to other global wildfire predictive models such as CLM and MC2.

\rev{While the present study introduces two fast surrogate models to emulate the process of \rev{JULES-INFERNO}, it is important to acknowledge certain limitations associated. Firstly, the surrogate models heavily rely on the assumption that the \rev{JULES-INFERNO} model accurately captures the complex dynamics of global wildfire prediction. Any limitations or uncertainties present in the JULES-INFERNO model~\cite{hantson2020quantitative} may propagate into the surrogate models' predictions. Additionally, due to the inherent complexity of wildfire prediction, it is challenging to precisely capture all the intricate spatial and temporal patterns solely through the surrogate models, especially when these patterns do not present in the training data. It is essential to carefully consider these limitations and their potential impact while interpreting and utilizing the results provided by the surrogate models in practical scenarios.}

Future work can be considered to surrogate the whole JULES system with more input, output variables, such as precipitation and humidity. The performance of these surrogate models can be enhanced when more training data and variables become available. This serves as a proof of concept by using deep learning with fine-tuning techniques to speed-up global wildfire prediction. 
It is reported that long-term predictions of JULES-INFERNO can suffer from model bias, compared to satellite observations~\cite{teckentrup2019response}. Due to the high computational cost, correcting the output of JULES-INFERNO in the full physical space can be challenging, especially when the \rev{observations} are partial and noisy.
Further efforts can be considered to apply latent data assimilation techniques~\cite{cheng_2022_datadriven} to efficiently adjust the surrogate model outputs using real-time observations during the online prediction phase.

\section*{Declaration of competing interest}
The authors declare that they have no known competing financial interests or personal relationships that could have appeared to influence the work reported in this paper.

\section*{Acknowledgements}
The authors would like to thank Andy Wiltshire and Chantelle Burton \rev{for providing output data from the JULES TRENDYv9 experiment, as well as providing invaluable advice and assistance on using the JULES-INFERNO model}. \rev{We thank the two anonymous reviewers for their careful reading of our manuscript and their many insightful comments and suggestions.}
 This research is funded by the Leverhulme Centre for Wildfires, Environment and Society through the Leverhulme Trust, grant number RC-2018-023. This work is partially supported by the French Agence Nationale de la Recherche (ANR) under reference ANR-22-CPJ2-0143-01. \rev{The} JULES-INFERNO \rev{historical (1990-2019) simulation was carried out by Andy Wiltshire at the UK Met Office. The four LGM} simulations were carried out using JASMIN, the UK's collaborative data analysis environment (https://jasmin.ac.uk).

\begin{acronym}[AAAAA]
\acro{NN}{Neural Networks}
\acro{RE}{\emph{Relative Error}}
\acro{SSIM}{\emph{Structural Similarity Index Measure}}
\acro{ML}{Machine Learning}
\acro{PR}{polynomial regression}
\acro{AE}{Auto-encoder}
\acro{AEP}{Absolute Error per Pixel}
\acro{CAE}{Convolutional Auto-encoder}
\acro{RNN}{Recurrent Neural Network}
\acro{CNN}{\emph{Convolutional Neural Network}}
\acro{LSTM}{\emph{Long Short-Term Memory}}
\acro{PCA}{Principal Component Analysis}
\acro{ROM}{reduced-order modelling}
\acro{CFD}{computational fluid dynamics}
\acro{1D}{one-dimensional}
\acro{2D}{two-dimensional}
\acro{3D}{three-dimensional}
\acro{NWP}{numerical weather prediction}
\acro{MSE}{Mean square error}
\acro{M21}{\emph{Many to One} strategy}
\acro{M2M}{\emph{Many to Many} strategy}
\acro{AI}{artificial intelligence}
\acro{DL}{deep learning}
\acro{ODE}{ordinary differential equations}
\acro{SVM}{Support Vector Machines}
\acro{ANN}{Artificial Neural Networks}
\acro{JULES-INFERNO}{JULES - INFERNO}
\acro{CA}{Cellular Automata}
\acro{RF}{Random Forests}
\acro{ESM}{Earth System Model}
\acro{LSM}{Land Surface Model}
\acro{INFERNO}{\emph{INteractive Fire and Emission algoRithm for Natural envirOnments}}
\acro{JULES}{\emph{Joint UK Land Environment Simulator}}
\acro{RH}{relative humidity}
\acro{DGVM}{\emph{dynamic global vegetation model}}
\acro{TRIFFID}{\emph{Top-down representation of interactive foliage and flora including dynamics}}
\acro{PFTs}{plant functional type}
\acro{ConvLSTM}{Convolutional LSTM}
\acro{CAE + LSTM}{CAE + LSTM}
\acro{best CAE - LSTM}{\emph{M2M Joint $12$ to $12$} CAE - LSTM}
\acro{best ConvLSTM}{\emph{M2M Joint $12$ to $12$} ConvLSTM}
\end{acronym}

\footnotesize{
\bibliographystyle{abbrv}
\bibliography{Bibliography}}

\end{document}